\documentclass[10pt,twocolumn,letterpaper]{article}
\usepackage[pagenumbers]{cvpr} 
\usepackage{graphicx}
\usepackage{amsmath}
\usepackage{amssymb}
\usepackage{booktabs}
\usepackage{inconsolata}
\usepackage{multirow}
\usepackage{marvosym}
\usepackage[pagebackref,breaklinks,colorlinks]{hyperref}
\usepackage[capitalize]{cleveref}
\crefname{section}{Sec.}{Secs.}
\Crefname{section}{Section}{Sections}
\Crefname{table}{Table}{Tables}
\crefname{table}{Tab.}{Tabs.}

\begin{document}
\title{Distilling Vision-Language Pre-training to Collaborate with \\ Weakly-Supervised Temporal Action Localization}

\author{Chen Ju\textsuperscript{1$\ast$}, \ \ Kunhao Zheng\textsuperscript{1$\ast$}, \ \ Jinxiang Liu\textsuperscript{1}, \ \ Peisen Zhao\textsuperscript{2}, \ \  Ya Zhang\textsuperscript{1\Letter}, \\
Jianlong Chang\textsuperscript{2}, \ \ Yanfeng Wang\textsuperscript{1}, \ \ Qi Tian\textsuperscript{2} 
\and
\textsuperscript{1}{Cooperative Medianet Innovation Center, Shanghai Jiao Tong University} \;    \textsuperscript{2}{Huawei Cloud \& AI}\\
{\tt\small \{ju\_chen,\;dyekuu,\;jinxliu,\;ya\_zhang,\;wangyanfeng\}@sjtu.edu.cn, \{pszhao93,\;jianlong.chang,\;tian.qi1\}@huawei.com}
}

\maketitle

\begin{abstract}
Weakly-supervised temporal action localization (WTAL) learns to detect and classify action instances with only category labels. Most methods widely adopt the off-the-shelf Classification-Based Pre-training (CBP) to generate video features for action localization. However, the different optimization objectives between classification and localization, make temporally localized results suffer from the serious incomplete issue. To tackle this issue without additional annotations, this paper considers to distill free action knowledge from Vision-Language Pre-training (VLP), since we surprisingly observe that the localization results of vanilla VLP have an over-complete issue, which is just complementary to the CBP results. To fuse such complementarity, we propose a novel distillation-collaboration framework with two branches acting as CBP and VLP respectively. The framework is optimized through a dual-branch alternate training strategy. Specifically, during the B step, we distill the confident background pseudo-labels from the CBP branch; while during the F step, the confident foreground pseudo-labels are distilled from the VLP branch. And as a result, the dual-branch complementarity is effectively fused to promote a strong alliance. Extensive experiments and ablation studies on THUMOS14 and ActivityNet1.2 reveal that our method significantly outperforms state-of-the-art methods. 
\end{abstract}

\vspace{-0.4cm}
\section{Introduction}   
Temporal action localization~(TAL), which aims to localize and classify action instances from untrimmed long videos, has been recognized as an indispensable component of video understanding~\cite{gao2017tall,yao2016highlight,shu2015joint}. To avoid laborious temporal boundary annotations, the weakly-supervised setting (WTAL)~\cite{wang2017untrimmednets,paul2018w,nguyen2018weakly,lee2020background}, {\em i.e.} only video-level category labels are available, has gained increasing attentions.

\begin{figure}[t]
\begin{center}
\includegraphics[width=0.478\textwidth] {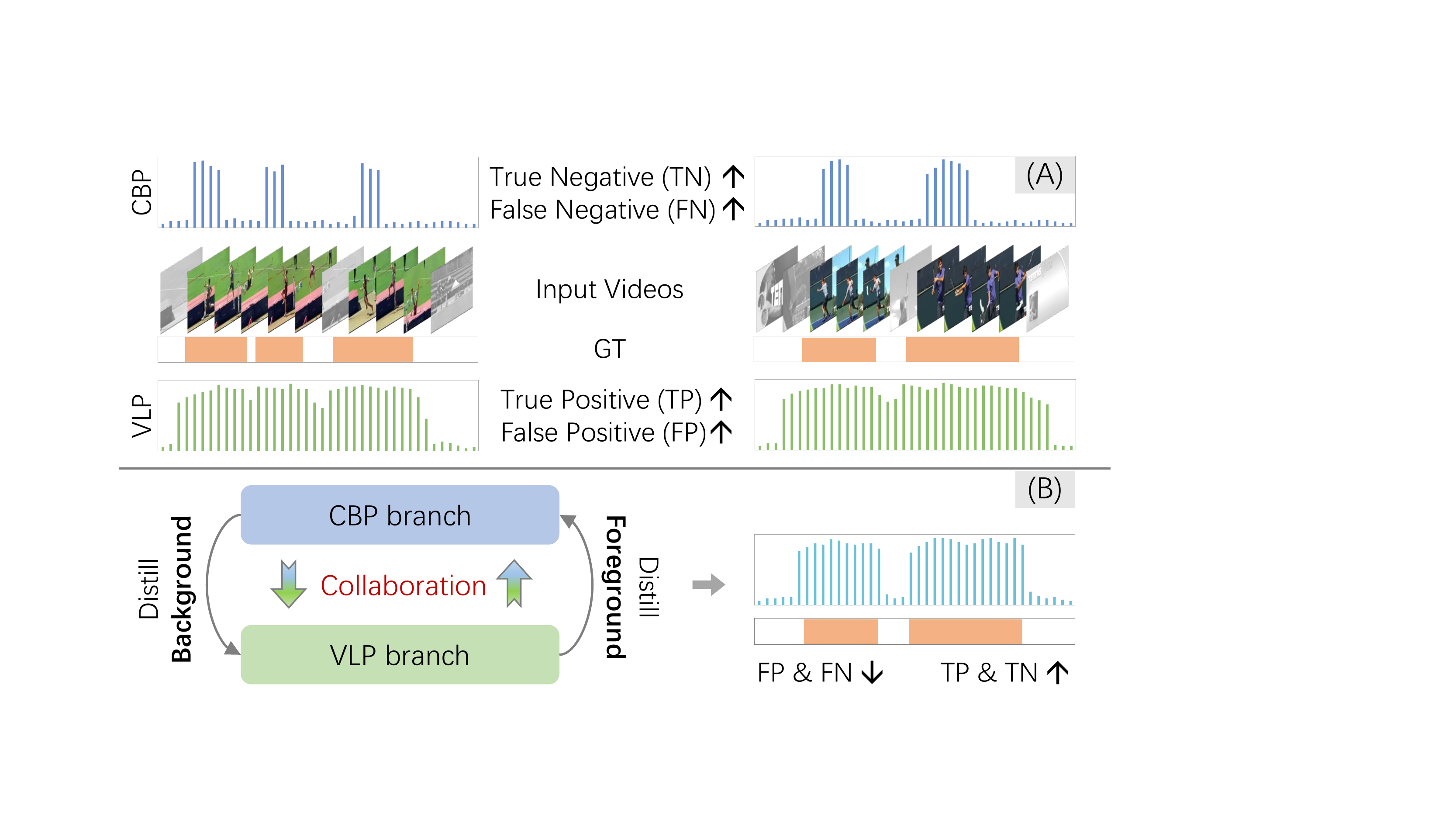} 
\end{center}
\vspace{-0.4cm}
\caption{\textbf{(A) Complementarity.} Most works use Classification-Based Pre-training (CBP) for localization, causing high TN yet serious FN. Vanilla Vision-Language Pre-training (VLP) confuses action and background, leading to high TP yet serious FP. \textbf{(B) Our distillation-collaboration framework} distills foreground from the VLP branch while background from the CBP branch, and promotes mutual collaboration, bringing satisfactory results.}
\label{fig:intro}
\end{figure}

To date in the literature, almost all WTAL methods rely on Classification-Based Pre-training~(CBP) for video feature extraction~\cite{simonyan2014two,carreira2017quo}. A popular pipeline is to train an action classifier with CBP features, then threshold the frame-level classification probabilities for final localization results. As demonstrated in Figure~\ref{fig:intro}\,(A), these CBP methods suffer from the serious \textbf{incompleteness issue}, {\em i.e.} only detecting sparse discriminative action frames and incurring \textit{high false negatives}. The main reason is that the optimization objective of classification pre-training, is to find several discriminative frames for action recognition, which is far from the objective of complete localization. As a result, features from CBP are inevitably biased towards partial discriminative frames. To solve the incompleteness issue, many efforts~\cite{min2020adversarial,liu2019completeness,lee2021weakly,zhai2020two} have been attempted, but most of them are trapped in a `performance-cost dilemma', namely, solely digging from barren category labels to keep costs low. Lacking location labels fundamentally limits the performance, leaving a huge gap from strong supervision.

To jump out of the dilemma, this paper raises one novel question: \textit{is there free action knowledge available, to help complete detection results while maintain cheap annotation overhead at the same time?} We naturally turn our sights to the prevalent Vision-Language Pre-training (VLP)~\cite{Radford21,Jia21}. VLP has demonstrated great success to learn joint visual-textual representations from large-scale web data. As language covers rich information about objects, human-object interactions, and object-object relationships, these learned representations could provide powerful human-object co-occurrence priors: valuable gifts for action localization.

We here take one step towards positively answering the question, {\em i.e.} fill the research gap of distilling action priors from VLP, namely CLIP~\cite{Radford21}, to solve the incomplete issue for WTAL. As illustrated in Figure\,\ref{fig:intro}\,(A), we first naively evaluate the temporal localization performance of VLP by frame-wise classification. But the results are far from satisfactory, suffering from the serious \textbf{over-complete issue}, {\em i.e.} confusing multiple action instances into a whole, causing \textit{high false positives}. We conjecture the main reasons are: (1) due to data and computational burden, almost all VLPs are trained using image-text pairs. Hence, VLP lacks sufficient temporal knowledge and relies more on human-object co-occurrence for localization, making it struggle to distinguish the actions with visually similar background contexts; (2) some background contexts have similar (confusing) textual semantics to actions, such as run-up {\em vs.} running.

Although simply steering VLP for WTAL is infeasible, we fortunately observe the \textbf{complementary property} between CBP and VLP paradigms: the former localizes high true negatives but serious false negatives, while the latter has high true positives but serious false positives. To leverage the complementarity, as shown in Figure\,\ref{fig:intro}\,(B), we design a novel distillation-collaboration framework that uses two branches to play the roles of CBP and VLP, respectively. The design rationale is to distill background knowledge from the CBP branch, while foreground knowledge from the VLP branch, for strong alliances. Specifically, we first warm up the CBP branch using only category supervision to initialize confident background frames, and then optimize the framework via an alternating strategy. \textit{During B step}, we distill background pseudo-labels for the VLP branch to solve the over-complete issue, hence obtaining high-quality foreground pseudo-labels. \textit{During F step}, we leverage high-quality pseudo-labels for the CBP branch to tackle the incomplete issue. Besides, in each step, we introduce both confident knowledge distillation and representation contrastive learning for pseudo-label denoising, effectively fusing complementarity for better results.

On two standard benchmarks: THUMOS14 and ActivityNet1.2, our method improves the average performance by 3.5\% and 2.7\% over state-of-the-art methods. We also conduct extensive ablation studies to reveal the effectiveness of each component, both quantitatively and qualitatively.

\begin{figure*}[t]
\begin{center}
\includegraphics[width=0.96\textwidth] {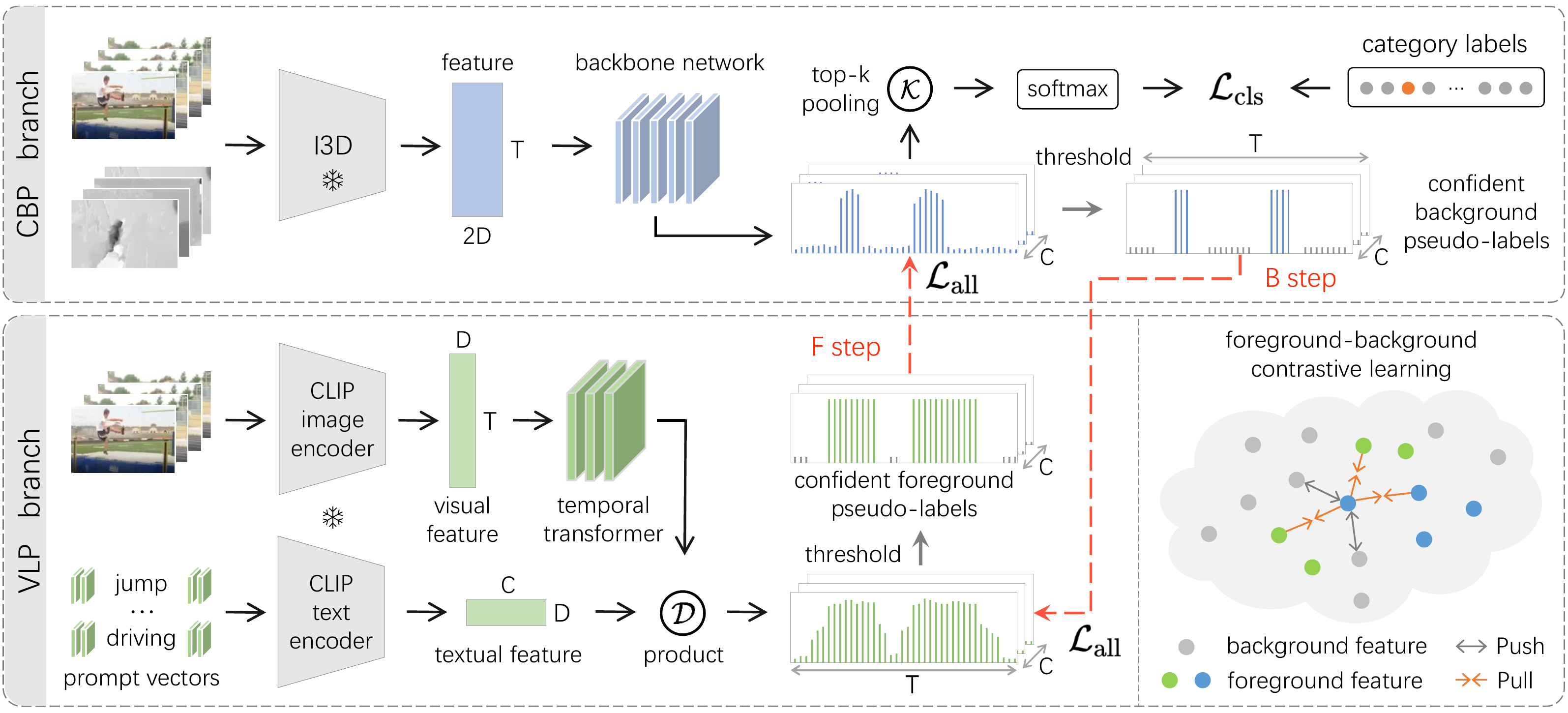}
\end{center}
\vspace{-0.4cm}
\caption{\small \textbf{Distillation-Collaboration Framework.} It covers two parallel branches, named CBP and VLP, and is optimized by an alternating strategy. We warm up the CBP branch in advance. In B step, we freeze both CLIP encoders, and distill confident background pseudo-labels from the CBP branch, to train prompt vectors and temporal Transformer in the VLP branch. In F step, confident foreground pseudo-labels are distilled for the CBP branch. We utilize both knowledge distillation loss and contrastive loss during dual-branch collaboration.}
\label{fig:framework}
\end{figure*}

To sum up, our contributions lie in three folds:

\vspace{0.05cm}
$\bullet$ We pioneer the first exploration in distilling free action knowledge from off-the-shelf VLP to facilitate WTAL;

\vspace{0.05cm}
$\bullet$ We design a novel distillation-collaboration framework that encourages the CBP branch and VLP branch to complement each other, by an alternating optimization strategy;

\vspace{0.05cm}
$\bullet$ We conduct extensive experiments and ablation studies to reveal the significance of distilling VLP and our superior performance on two public benchmarks.

\section{Related Work}
\noindent \textbf{Vision-Language Pre-training (VLP)} aims to learn cross-modal representations~\cite{Mori99,Weston11,frome2013devise} from large-scale web data. Comparing to video, image requires fewer costs for annotation and computation, hence almost all VLPs are image-based, {\em e.g.}~\cite{Radford21,Jia21,yao2021filip,yu2022coca,alayrac2022flamingo,wang2022ofa}. Recently, several studies adopted VLP to provide free visual-semantic knowledge for downstream image tasks, such as detection~\cite{gu2021open,zareian2021open}, segmentation~\cite{rao2021denseclip,zhou2022extract}, human-object interaction~\cite{khandelwal2021simple,liao2022gen}, synthesis~\cite{liu2021more}, and generation~\cite{wang2022clip,crowson2022vqgan,tevet2022motionclip}. In terms of the video domain, \cite{luo2021clip4clip,wang2021actionclip,lin2022frozen} equipped VLP with temporal transformers for action recognition. \cite{ju2021prompting,nag2022zero} introduced prompt learning for efficient retrieval or detection. However, these studies focus more on open-vocabulary scenarios or short video understanding. On the contrary, this paper makes the first exploration to steer VLP for long video temporal localization, under the weakly-supervised setting.

\vspace{0.1cm}
\noindent \textbf{Strongly-supervised Temporal Action Localization} has achieved great progress~\cite{zeng2019graph,liu2022empirical,lin2021learning,qing2021temporal}, given precise action boundaries and categories. There are two popular pipelines: the top-down framework~\cite{shou2016temporal,shou2017cdc,gao2017turn,chao2018rethinking,lin2017single,xu2017r,xu2020g,tan2021relaxed,zhu2021enriching,wang2022rcl} pre-defines massive anchors based on the action distribution prior, and uses fixed-length sliding windows to generate initial proposals, then regresses to refine boundaries; the bottom-up framework~\cite{zhao2017temporal,lin2018bsn,lin2019bmn,lin2019fast,zhao2020bottom,bai2020boundary,xu2020g,nag2022temporal,yang2022temporal,su2021bsn++,cheng2022tallformer} trains frame-wise boundary detectors for extreme frames (start, end, center), then groups extreme frames or estimates action lengths to produce proposals. In addition, some papers~\cite{gao2018ctap,liu2019multi,yang2020revisiting} proposed various fusion strategies to complement these frameworks. Several studies~\cite{zeng2019graph,liu2020progressive} are devoted to better post-processing. Nevertheless, all the above methods demand precise boundary annotations, which are time-consuming and expensive in reality.

\vspace{0.1cm}
\noindent \textbf{Weakly-supervised Temporal Action Localization} significantly alleviates annotation costs, training with only category labels. The pivotal component is CAS~\cite{wang2017untrimmednets,paul2018w,nguyen2018weakly,islam2021hybrid} obtained from Classification-Based Pre-training (CBP). But due to the gap of classification and localization, CBP suffers from the serious incomplete issue: only detect discriminative action fragments or even background. To solve this issue, \cite{zhong2018step,min2020adversarial,ju2021adaptive} introduced the erasing strategy. \cite{liu2019completeness,luo2021action,narayan2021d2} produced multiple CAS in parallel for complementarity.~\cite{nguyen2019weakly,lee2019background,lee2021weakly,shi2020weakly} proposed background modeling and context separation.~\cite{gao2022fine,he2022asm,huang2021foreground,zhang2021cola} enhanced features by intra- and inter-video modeling. To iteratively refine results, some recent papers~\cite{luo2020weakly,zhai2020two,yang2021uncertainty} introduced the self-training strategy, \cite{shou2018autoloc,liu2019weakly} adopted outer-inner contrastive learning. While encouraging, most of them are trapped in the `performance-cost dilemma', {\em i.e.} solely digging from barren category labels to keep costs low. But the lack of annotations leads to a huge performance gap between strong and weak supervisions. To this end,~\cite{xu2018segregated,narayan20193c,ma2020sf,ju2020point,ju2021divide,lee2021learning} explored the idea of adding instance-number or single-frame annotations for further improvements. As a fresh departure from existing work, to pursue better performance without additional annotation costs, this paper intends to distill free action knowledge from off-the-shelf VLPs, to assist WTAL.

\section{Method}
\subsection{Notations and Preliminaries} 
\noindent \textbf{Task Formulation.}  
Given $N$ untrimmed videos $\{{v_i}\}_{i=1}^N$, and their video-level category labels $\{\mathbf{y}_i \in\mathbb{R}^{C}\}_{i=1}^N$, where $C$ means the total number of action categories, WTAL intends to detect and classify all action instances, in terms of a set of quadruples $\{(s,e,c,p)\}$, where $s$, $e$, $c$, $p$ represent the start time, the end time, the action category and detection score of the action proposal, respectively. Note that each video may contain multiple action instances.

\vspace{0.1cm}
\noindent \textbf{Motivation.}
There exists significant complementarity between the localization results of Classification-Based Pre-training and Vision-Language Pre-training, as concluded in Table~\ref{tab:complementarity}. The former suffers from incomplete results (serious false negatives), but has good true negatives; the latter suffers from over-complete results (serious false positives), but has good true positives. Such investigations motivate us to collaborate the complementary results for strong alliances, through a distillation-collaboration framework.

\vspace{0.1cm}
\noindent \textbf{Framework Overview.}
As demonstrated in Figure~\ref{fig:framework}, the distillation-collaboration framework consists of two parallel branches, named CBP branch and VLP branch respectively, and is optimized by an alternating strategy. The CBP branch is first warmed-up to produce rich background information with only classification supervision. During B step, we distill confident background pseudo-labels from CBP branch, for VLP branch to tackle the over-complete issue, and thus localize high-quality foreground and background information. During F step, we distill superior pseudo-labels from the well-trained VLP branch, for CBP branch to tackle the incomplete issue. Through such dual-branch collaboration, we effectively fuse the complementary results.

\subsection{Foreground and Background Distillation}
\noindent \textbf{The CBP Branch} is utilized to identify a large number of background frames as well as several discriminative action frames, by exploiting Classification-Based Pre-training.

Following literature~\cite{liu2019completeness,lee2021weakly}, we adopt CB pre-training, {\em i.e.} the I3D architecture pre-trained on Kinetics~\cite{carreira2017quo}, to extract RGB and Flow features, and then concatenate them to form the two-stream features $\mathbf{F}_{\mathrm{i3d}} \in \mathbb{R}^{T \times 2D}$, where $T$ and $D$ refer to the temporal length and feature dimension. Next, feeding with $\mathbf{F}_{\mathrm{i3d}}$, the CBP branch uses the backbone network for feature fine-tuning and localization, and finally outputs the frame-level action probabilities $\mathbf{P}^{\mathrm{cb}} \in \mathbb{R}^{T \times C}$.

To achieve action classification, we adopt multiple instance learning, {\em i.e.} for the output $\mathbf{P}^{\mathrm{cb}}$ from the backbone network, we aggregate (pool) the top-$k$ frames' scores as video-level category scores $\mathbf{\widehat{y}} \in \mathbb{R}^{C}$, then supervise it via the binary cross-entropy loss, which is formulated as:
\begin{equation}
    {\mathcal{L}_{\mathrm{cls}} = \sum_{c=1}^{C} -y_{c}\; \mathrm{log}\;\widehat{y}_{c},
    \quad
    \widehat{y}_{c} = \sigma(\frac{1}{k} \sum \mathcal{K}(\mathbf{P}^{\mathrm{cb}})),}
\end{equation}
where $\mathcal{K}$ denotes the top-$k$ score set in the temporal domain, and $\sigma$ refers to the softmax function.

\vspace{0.15cm}
\noindent \underline{\textit{Remark.}} Under CB Pre-training and category-only supervision, $\mathbf{P}^{\mathrm{cb}}$ is well known for focusing on sparse discriminative action frames, {\em i.e.} high true negatives but serious false negatives, thus can provide rich background information.

\vspace{0.2cm}
\noindent \textbf{The VLP Branch} is designed to mine free action knowledge from VL Pre-training, {\em e.g.} CLIP~\cite{Radford21}. Since image-text pre-training lacks sufficient temporal priors, its vanilla localization results have a serious over-complete issue, {\em i.e.} high false positives. To tackle this issue, we propose to fine-tune CLIP using extensive background samples.

Instead of linear probing, we use efficient prompt learning~\cite{lester2021power,ju2021prompting} for fine-tuning: \textit{we freeze the CLIP backbone, only optimize several prompt vectors and temporal layers}.

Concretely, for the visual stream, we first split the video into consecutive frames, and then utilize the CLIP image encoder to extract frame-level features $\mathbf{F}_{\mathrm{vis}} \in \mathbb{R}^{T \times D}$. For temporal relationship construction, we strengthen $\mathbf{F}_{\mathrm{vis}}$ into $\mathbf{F}_{\mathrm{vid}} \in \mathbb{R}^{T \times D}$ through simple temporal transformer layers $\Phi_{\mathrm{temp}}(\cdot)$. While for the textual stream, we first prepend and append several learnable prompt vectors $\Phi_{\mathrm{prmp}}(\cdot)$ to category names, then feed them into the CLIP text encoder, to obtain textual features $\mathbf{F}_{\mathrm{txt}} \in \mathbb{R}^{C \times D}$. 
Formally,
\begin{equation}
    {\mathbf{F}_{\mathrm{vid}} = \Phi_{\mathrm{temp}}(\mathbf{F}_{\mathrm{vis}}), 
    \ \  
    \mathbf{F}_{\mathrm{txt}} = \Phi_{\mathrm{txt}}(\Phi_{\mathrm{prmp}}(\mathrm{C}_{\mathrm{name}})),
    }
\end{equation}
where $\mathrm{C}_{\mathrm{name}}$ refers to action category names, and $\Phi_{\mathrm{txt}}(\cdot)$ is the CLIP textual encoder. Thereafter, the frame-level localization results $\mathbf{P}^{\mathrm{vl}}$ for this branch can be calculated as:
\begin{equation}
    {\mathbf{P}^{\mathrm{vl}} = \sigma(\mathbf{F}_{\mathrm{vid}} \cdot \mathbf{F}^{\mathsf{T}}_{\mathrm{txt}}) \in \mathbb{R}^{T \times C}}.
\end{equation}

\noindent \underline{\textit{Remark.}} \hspace{0.5pt} For the VLP branch, we only optimize lightweight model parameters for false-positive suppression, naturally bringing two main benefits: (1) the frozen CLIP backbone preserves the action prior knowledge in pre-training, thus maintaining high true-positive results; (2) it matches the demand for less supervision data under weakly-supervised settings, and also saves the memory footprint.

\vspace{0.1cm}
\noindent \textbf{Confident Pseudo-labels.}
\hspace{0.5pt} Since both $\mathbf{P}^{\mathrm{vl}}$ and $\mathbf{P}^{\mathrm{cb}}$ contain somewhat noise, to make reliable use of complementary information, we distill confident location pseudo-labels of foreground and background respectively. That is, for the CBP branch, we distill extensive background pseudo-labels from the output $\mathbf{P^{\mathrm{cb}}}$; while for the VLP branch, we distill sufficient foreground pseudo-labels from the output $\mathbf{P^{\mathrm{vl}}}$.

For both branches, we leverage double thresholds $\delta_{h}$ and $\delta_{l}$ $(\delta_{h} > \delta_{l})$, to convert localization results $\mathbf{P}$ into \textit{ternary} pseudo-labels $\mathbf{H} \in \mathbb{R}^{T \times C}$, which are formally written as:
\begin{equation}
h_{t,c}=\left\{
\begin{aligned}
1 \ \ & \mathrm{if} \ \ p_{t} > \delta_{h} \ \ \mathrm{and} \ \ p_c = y_c \\
0 \ \ & \mathrm{if} \ \ p_{t} < \delta_{l} \quad  \mathrm{or} \quad p_c \neq y_c  \\
-1 \ \ & \qquad \quad \ \  \mathrm{otherwise} 
\end{aligned}
\right.
\end{equation}
where the subscripts $c$ and $t$ refer to the indices of category and frame. More specifically, for either branch, we regard frames with scores more than $\delta_{h}$ and the correct action category as the foreground; while frames with scores less than $\delta_{l}$ or with the wrong action category are treated as the background; the remaining frames are considered uncertain. As a result, the pseudo-labels $\mathbf{H}^{\mathrm{cb}}$ cover vast confident background frames, as well as trivial foreground frames; while the pseudo-labels $\mathbf{H}^{\mathrm{vl}}$ contain dense confident foreground frames, and partial background frames. Note that, for both branches, we generate positive and negative frames to avoid trivial solutions, and facilitate contrastive learning for feature enhancement, as detailed in the following section.

\subsection{Dual-Branch Collaborative Optimization}
In this section, we encourage two branches to collaborate with each other, such that forming a strong alliance of their complementary localization results. To reduce serious noises in pseudo-labels, we introduce an alternate training strategy for dual-branch collaborative optimization.

The design rationale is to distill \textbf{B}ackground knowledge from the CBP branch for \textbf{B} step, while distill \textbf{F}oreground knowledge from the VLP branch in \textbf{F} step. To be specific, we warm up the CBP branch in advance, using only category supervision, to initialize reliable background frames. During B Step, we freeze the well-trained CBP branch, and then generate confident background pseudo-labels $\mathbf{H}^{\mathrm{cb}}$ to supervise the VLP branch. As a result, these false-positive confusions from vanilla CLIP pre-training are greatly tackled, and the resultant pseudo-labels contain a large number of confident foreground frames and background frames. During F Step, the high-quality pseudo-labels $\mathbf{H}^{\mathrm{vl}}$ are distilled from the frozen VLP branch, to guide the CBP branch for the false-negative suppression. Under such an alternating strategy, these two branches not only complement each other, but also correct each other, thus jointly contributing to more precise and complete action localization.

During each step, to supervise either branch, we adopt both the knowledge distillation loss $\mathcal{L}_{\mathrm{kd}}$ and foreground-background contrastive loss $\mathcal{L}_{\mathrm{fb}}$. The total optimization loss can be written with a balancing ratio $\lambda$, as follows:
\begin{equation}
    {\mathcal{L}_{\mathrm{all}} = \mathcal{L}_{\mathrm{kd}}(\mathbf{H'}, \mathbf{P}) \; + \;
    \lambda \mathcal{L}_{\mathrm{fb}}(\Psi^{\mathrm{+}}, \Psi^{\mathrm{-}})}.
\end{equation}

Here, $\mathcal{L}_{\mathrm{kd}}$ regularizes one branch to output similar detection results with pseudo-labels from the other branch. Given that there exist some noises in pseudo-labels, {\em i.e.} uncertain frames, we only make supervision on confident frames.
\begin{equation}
    {\mathcal{L}_{\mathrm{kd}}(\mathbf{H'}, \mathbf{P}) = \frac{1}{O}\sum_{c=1}^{C} \sum_{t=1}^{O} 
    D_{\mathrm{KL}}(h'_{t,c} \, || \, p_{t,c}),}
\end{equation}
where $D_{\mathrm{KL}}(p(x)\,||\,q(x))$ refers to the Kullback-Leibler divergence of distribution $p(x)$ from distribution $q(x)$, $O$ is the total number of confident frames, and $\mathbf{H'}$ denotes the pseudo-labels from the other branch. Note that, the pseudo-labels contain two types of confident frames: foreground and background, which could help to avoid falling into the trivial solutions under one single type of labels.

Moreover, in untrimmed long videos, some background contexts could appear visually similar with the action (foreground). We further introduce contrastive learning to pull foreground features and push background features. Specifically, we treat confident foreground frames from the same action category as the positive set $\Psi^{\mathrm{+}}$, while all confident background frames as the negative set $\Psi^{\mathrm{-}}$, then foreground-background contrastive loss is formulated as:
\begin{align}
\mathcal{L}_{\mathrm{fb}}(\Psi^{\mathrm{+}}_{i}, \Psi^{\mathrm{-}}_{i}) = & \sum_i - \log \frac{\sum_{m \in {\Psi^{\mathrm{+}}_{i}}}  \exp(\mathbf{f}_i \cdot \mathbf{f}_{m}/ \tau)}{\sum_{j \in {*}} \exp(\mathbf{f}_i \cdot \mathbf{f}_{j}/ \tau)},
\end{align}
where $\mathbf{f} \in \mathbb{R}^{D}$ refers to the frame feature, $\tau$ means the temperature hyper-parameter for scaling, and $*$ means the union of $\Psi^{\mathrm{+}}_{i}$ and $\Psi^{\mathrm{-}}_{i}$. Take the VLP branch as an example, sufficient background pseudo-labels contain considerable hard negative samples, to help distinguish features of foreground and contexts. In addition, the enhanced features of uncertain frames also become more discriminative, further facilitating more complete temporal action localization.

\begin{table*}[t]
\small
\centering
\setlength\tabcolsep{7pt}
\begin{tabular}{c|c|c|ccccccccc}
\toprule
 &  &  & \multicolumn{7}{c|}{mAP@IoU} &  &  \\ \cline{4-10}
\multicolumn{1}{c|}{\multirow{-2}{*}{Supervision}} & \multicolumn{1}{c|}{\multirow{-2}{*}{Method}} & \multirow{-2}{*}{Feature} &  0.1 & 0.2 & 0.3 & 0.4 & 0.5 & 0.6 & \multicolumn{1}{c|}{0.7} & \multirow{-2}{*}{\begin{tabular}[c]{@{}c@{}}AVG\\(0.1-0.5)\end{tabular}} & \multirow{-2}{*}{\begin{tabular}[c]{@{}c@{}}AVG\\(0.3-0.7)\end{tabular}} \\ \hline \hline
\multirow{5}{*}{Strong} & DBS~\cite{gao2019video} & \multirow{5}{*}{I3D} & 56.7 & 54.7 & 50.6 & 43.1 & 34.3 & 24.4 & \multicolumn{1}{c|}{14.7} & 47.9 & 33.4 \\
& BUMR~\cite{zhao2020bottom} &  & - & - & 53.9 & 50.7 & 45.4 & 38.0 & \multicolumn{1}{c|}{28.5} & - & 43.3 \\
& GCM~\cite{zeng2019graph} &  & 72.5 & 70.9 & 66.5 & 60.8 & 51.9 & - & \multicolumn{1}{c|}{-} & \textbf{64.5} & - \\
& RCL~\cite{wang2022rcl} &  & - & - & 70.1 & 62.3 & 52.9 & 42.7 & \multicolumn{1}{c|}{30.7} & - & 51.7 \\
& ActionFormer~\cite{zhang2022actionformer} &  & - & - & 82.1 & 77.8 & 71.0 & 59.4 & \multicolumn{1}{c|}{43.9} & - & \textbf{66.8} \\
\hline \hline
\multirow{15}{*}{Weak} 
& HAM-Net~\cite{islam2021hybrid} & \multirow{10}{*}{I3D} & 65.4 & 59.0 & 50.3 & 41.1 & 31.0 & 20.7 & \multicolumn{1}{c|}{11.4} & 49.4 & 30.9 \\
& UM~\cite{lee2021weakly} & & 63.0 & 56.6 & 49.0 & 40.9 & 30.4 & 21.0 & \multicolumn{1}{c|}{10.4} & 48.0 & 30.3 \\
& TS-PCA~\cite{liu2021blessings} & & 67.6 & 61.1 & 53.4 & 43.4 & 34.3 & 24.7 & \multicolumn{1}{c|}{13.7} & 52.0 & 33.9 \\
& FTCL~\cite{gao2022fine} & & 69.6 & 63.4 & 55.2 & 45.2 & 35.6 & 23.7 & \multicolumn{1}{c|}{12.2} & 53.8 & 34.4 \\
& ACGNet~\cite{yang2021acgnet} & & 68.1 & 62.6 & 53.1 & 44.6 & 34.7 & 22.6 & \multicolumn{1}{c|}{12.0} & 52.6 & 33.4 \\
& DCC~\cite{li2022exploring} & & 69.0 & 63.8 & 55.9 & 45.9 & 35.7 & 24.3 & \multicolumn{1}{c|}{13.7} & 54.1 & 35.1 \\
& ASM-Loc~\cite{he2022asm} & & 71.2 & 65.5 & 57.1 & 46.8 & 36.6 & 25.2 & \multicolumn{1}{c|}{13.4} & 55.4 & 35.8 \\
& CO2-Net~\cite{hong2021cross} & & 70.1 & 63.6 & 54.5 & 45.7 & 38.3 & 26.4 & \multicolumn{1}{c|}{13.4} & 54.4 & 35.7 \\
& RSKP~\cite{huang2022weakly} & & 71.3 & 65.3 & 55.8 & 47.5 & 38.2 & 25.4 & \multicolumn{1}{c|}{12.5} & 55.6 & 35.9 \\
& DELU~\cite{chen2022dual} & & 71.5 & 66.2 & 56.5 & 47.7 & 40.5 & 27.2 & \multicolumn{1}{c|}{15.3} & 56.5 & 37.4 \\
\cline{2-12}
& CO2-Net$\dag$ & \multirow{5}{*}{\begin{tabular}[c]{@{}c@{}}I3D\\ + \\CLIP\end{tabular}} & 66.5 & 59.4 & 50.7 & 41.7 & 34.2 & 22.4 & \multicolumn{1}{c|}{12.0} & 50.5 & 32.2 \\
& CO2-Net$\ddag$ &  &  68.8 & 62.0 & 51.7 & 42.2 & 35.4 & 22.3 & \multicolumn{1}{c|}{11.7} & 51.9 & 32.7 \\
& DELU$\dag$ & & 68.5 & 61.2 & 52.1 & 43.1 & 35.0 & 23.1 & \multicolumn{1}{c|}{12.7} & 52.0 & 33.2 \\
& DELU$\ddag$ & & 70.5 & 64.5 & 55.2 & 45.7 & 38.5 & 25.7 & \multicolumn{1}{c|}{13.8} & 54.9 & 35.8 \\
& \textbf{Ours} &  & \textbf{73.5} & \textbf{68.8} & \textbf{61.5} & \textbf{53.8} & \textbf{42.0} & \textbf{29.4} & \multicolumn{1}{c|}{\textbf{16.8}} & \textbf{60.0} & \textbf{40.8} \\
\bottomrule
\end{tabular}
\vspace{-0.1cm}
\caption{\textbf{Comparison with state-of-the-art methods on THUMOS14.} For fair comparisons, we reproduce the results of SOTA methods: CO2-Net~\cite{hong2021cross} and DELU~\cite{chen2022dual}, by inputting both I3D~\cite{carreira2017quo} and CLIP~\cite{Radford21} features. $\dag$ and $\ddag$ refer to averaging or concatenating these two features. AVG(0.1-0.5) and AVG(0.3-0.7) are the average mAP from IoU 0.1 to 0.5 and from IoU 0.3 to 0.7. Our framework significantly surpasses all weakly-supervised competitors using identical features, and is even comparable to early strongly-supervised methods.}
\vspace{0.01cm}
\label{tab:THUMOS}
\end{table*}

\begin{table}[t]
\small
\centering
\vspace{0.2cm}
\setlength\tabcolsep{7.5pt}
\begin{tabular}{c|c|ccc|c}
\toprule
\multicolumn{1}{c|}{\multirow{2}{*}{Method}} & \multicolumn{1}{c|}{\multirow{2}{*}{Feature}} & \multicolumn{3}{c|}{mAP@IoU} & \multirow{2}{*}{AVG} \\ \cline{3-5}
 & & 0.5 & 0.75 & 0.95 &  \\ \hline \hline
CleanNet~\cite{liu2019weakly} & \multirow{10}{*}{I3D} & 37.1 & 20.3 & 5.0 & 21.6 \\
CMCS~\cite{liu2019completeness} & & 36.8 & 22.0 & 5.6 & 22.4 \\
TSCN~\cite{zhai2020two} & & 37.6 & 23.7 & 5.7 & 23.6 \\ 
BasNet~\cite{lee2020background} & & 38.5 & 24.2 & 5.6 & 24.3 \\
DGAM~\cite{shi2020weakly} & & 41.0 & 23.5 & 5.3 & 24.4 \\ 
UM~\cite{lee2021weakly} & & 41.2 & 25.6 & 6.0 & 25.9 \\
ACGNet~\cite{yang2021acgnet} & & 41.8 & 26.0 & 5.9 & 26.1 \\ 
D2-Net~\cite{narayan2021d2} & & 42.3 & 25.5 & 5.8 & 26.0 \\ 
CO2-Net~\cite{hong2021cross} & & 43.3 & 26.3 & 5.2 & 26.4 \\
DELU~\cite{chen2022dual} & & 44.2 & 26.7 & 5.4 & 26.9 \\  \hline
CO2-Net$\dag$ & \multirow{5}{*}{\begin{tabular}[c]{@{}c@{}}I3D\\ + \\CLIP\end{tabular}} & 44.3 & 26.6 & 5.4 & 26.9 \\
CO2-Net$\ddag$ &  & 44.7 & 26.9 & 5.8 & 27.4 \\
DELU$\dag$ & & 44.9 & 26.9 & 5.6 & 27.2 \\
DELU$\ddag$ &  & 45.6 & 27.5 & 5.8 & 27.8 \\
\textbf{Ours} & & \textbf{48.3} & \textbf{29.3} & \textbf{6.1} & \textbf{29.6} \\ \bottomrule
\end{tabular}
\vspace{-0.1cm}
\caption{\textbf{Comparison with state-of-the-art methods on ActivityNet1.2.} For fair comparisons, we reproduce CO2-Net~\cite{hong2021cross} and DELU~\cite{chen2022dual}, by inputting I3D~\cite{carreira2017quo} and CLIP~\cite{Radford21} features. $\dag$ and $\ddag$ refer to averaging or concatenating these features. AVG is the average mAP at the thresholds 0.5:0.05:0.95. Our method significantly surpasses all competitors, especially at loose IoU thresholds.}
\label{tab:ANET}
\end{table}

\vspace{0.1cm}
\noindent \underline{\textit{Discussion.}} 
\hspace{0.5pt} Comparing to various fusion strategies, our alternating strategy produces more precise and complete results (see Table~\ref{tab:strategy} for details). This strategy plays a similar role as the multi-view co-training, where CBP and VLP branches can be deemed as two distinctive views, thus being quite robust to pseudo-label noises. For B step, we prompt the VLP branch for high-quality pseudo-labels, where some frames with conflicting predictions are still treated as uncertain. In F step, we use feature contrastive loss to make their results more discriminative, {\em i.e.} further denoising.

\subsection{Inference}
At testing time, we leverage the results from the CBP branch for post-processing, as vision-language pre-training cannot handle Optical Flow, which is essential for WTAL. Given an input video, we first obtain video-level category probabilities and frame-level localization scores. For action classification, we select the classes with probability greater than $\theta_{cls}$; and for localization, we threshold detection scores with $\theta_{loc}$, concatenate consecutive snippets as action proposals, and eliminate redundant proposals with soft non-maximum suppression (NMS). Each proposal is scored with the detection maximum in the proposal interval.

\section{Experiments}
\subsection{Implementation}
\noindent \textbf{Datasets.} \textbf{THUMOS14} owns 413 untrimmed videos from 20 categories, and each video contains an average of 15 instances. As conventions, we train on 200 validation videos, and evaluate on 213 testing videos. Despite its small scale, this dataset is challenging, since video lengths vary widely and actions occur frequently.
\noindent \textbf{ActivityNet1.2} covers 9682 videos of 100 categories, dividing into 4619 training videos, 2383 validation videos, and 2480 testing videos. Almost all videos contain one single category, and action regions take up more than half of the duration in most videos. We train on the training set and evaluate on the validation set.

\vspace{0.1cm}
\noindent \textbf{Metrics.} To evaluate localization performance, we follow the standard protocol to use mean Average Precision (mAP) at different intersections over union (IoU) thresholds. Note that a proposal is regarded as positive only if both the category prediction is correct and IoU exceeds the set threshold. To clearly evaluate the quality of pseudo-labels, we also report mean Intersection over Union (mIoU) averaged over the foreground categories and the background category.

\vspace{0.1cm}
\noindent \textbf{Details.}
To handle the large variety in video durations, we randomly sample $T$ consecutive snippets for each video. $T$ is set to $1000$ on THUMOS14, and $400$ on ActivityNet1.2. We utilize the TV-L1 algorithm to extract optical flow from RGB data. For the CBP branch, we use Transformer architectures (multi-head self-attention, layer norm, and MLPs) as the backbone network. For the VLP branch, we use a $2$-layer temporal Transformer, prepend and append $16$ prompt vectors to textual inputs, both are initialized by $\mathcal{N}(0,0.01)$. Both CLIP image encoder and text encoder are adopted from ViT-B/16. The framework is optimized by Adam with the learning rate of $10^{-4}$. All hyper-parameters are set by grid search: pseudo-label thresholds $\delta_{h}=0.3$, $\delta_{l}=0.1$, inference thresholds $\theta_{cls}=0.85$, $\theta_{loc}=0.45$, the balancing ratio $\lambda=0.05$, the temperature $\tau=0.07$.

\subsection{Comparison with State-of-the-art Methods}
Here, we make comprehensive comparisons with current state-of-the-art methods across multiple IoU thresholds.

\vspace{0.05cm}
\noindent \textbf{Performance.}
\hspace{1pt} The comparisons on THUMOS14 are provided in Table~\ref{tab:THUMOS}. Here, we separate two levels of supervision: strong and weak, for better quantification. Generally speaking, our framework achieves new state-of-the-art on all IoU regimes. Comparing with recent methods, the gain of the average mAP (0.3-0.7) even reaches 4-5\%, further narrowing the performance gap between weak and strong supervisions. Moreover, our method achieves considerable gains on strict and loose evaluations. For example, when comparing to DELU~\cite{chen2022dual}, the gains are 3.5\% average mAP (0.1-0.5) and 3.4\% average mAP (0.3-0.7), indicating that our results are complete and precise. Furthermore, despite being weakly-supervised settings, at several low IoU thresholds, our framework even performs comparably with some earlier strongly-supervised methods~\cite{gao2019video,gao2017turn,zhao2017temporal}.

Table~\ref{tab:ANET} shows the comparison results on ActivityNet1.2. On all IoU thresholds, our designed framework surpasses existing methods by a large margin. In terms of the average mAP, the performance improvement can reach 2.7\%, taking the state-of-the-art to a new level. However, due to the lack of precise location annotations, the gain decreases as the IoU threshold becomes stricter, {\em e.g.} 4.1\%\,@IoU\,0.5 {\em vs.} 0.7\%\,@IoU\,0.95, when comparing to DELU~\cite{chen2022dual}.

\noindent \textbf{Source of Gain.}
Comparing to existing methods, we leverage Vision-Language Pre-training for free knowledge. To make fair comparisons, we also input both I3D and CLIP features into two SOTA methods~\cite{hong2021cross,chen2022dual}, by one early fusion mode (average or concatenate these features). In general, simply adding CLIP features gives only slight or even negative gains on both datasets, which is due to the bad over-complete issue of CLIP (detailed in Table~\ref{tab:complementarity}). For THUMOS with complex and frequent actions, over-complete worsens results. While in ActivityNet, most videos only contain one action that takes up half of the video duration, thus benefiting a bit from over-complete. Nevertheless, with identical features, our method significantly outperforms existing competitors, proving the effectiveness of our framework.

\begin{table}[t]
\small
\centering
\setlength\tabcolsep{10.6pt}
\begin{tabular}{c|cc|cc}
\toprule    
\multirow{2}{*}{Setting} & \multicolumn{2}{c|}{THUMOS} & \multicolumn{2}{c}{ActivityNet} \\ \cline{2-5} & Fore & Back & Fore & Back \\ \hline  \hline
CB Pre-training & 56.0 & 88.7 & 54.7 & 88.3  \\ 
VL Pre-training & 75.6 & 34.1 & 72.1 & 45.0 \\  
Ours & 72.4 & 80.0 & 69.5 & 71.7 \\
\bottomrule
\end{tabular}
\vspace{-0.1cm}
\caption{\textbf{Complementarity of pre-training.} CB Pre-training has good background mIoU but inferior foreground mIoU. VL Pre-training has good foreground mIoU but inferior background mIoU. Our method achieves both high foreground and background mIoU.}
\vspace{0.5cm}
\label{tab:complementarity}
\end{table}

\begin{table}[t]
\small
\centering
\setlength\tabcolsep{7.5pt}
\begin{tabular}{c|ccc|c}
\toprule
\multirow{2}{*}{Fusion Strategy} & \multicolumn{3}{c|}{mAP@IoU} & \multirow{2}{*}{\begin{tabular}[c]{@{}c@{}}AVG\\ (0.3-0.7)\end{tabular}} \\ \cline{2-4}
 & 0.3 & 0.5 & 0.7 & \\ \hline \hline
Only F step \ (AVG) & 40.0 & 16.1 & 3.3 & 19.1 \\
Only F step \ (Weight) & 52.8  & 25.5 & 6.1 & 27.7 \\ \hline
Only B step \ (AVG) & 56.7 & 33.6 & 11.3 & 34.1 \\
Only B step \ (Weight) & 58.5  & 38.8 & 14.8 & 37.8 \\  \hline
Alternating & \textbf{61.5} & \textbf{42.0} & \textbf{16.8} & \textbf{40.8} \\ 
\bottomrule
\end{tabular}
\vspace{-0.1cm}
\caption{\textbf{Comparison of optimization strategies.} ``Only F step'': extract pseudo-labels from vanilla VLP to train the CBP branch. ``Only B step'': get pseudo-labels from the warm-up CBP branch to train the VLP branch. We combine two branches by averaging and weighting. Our alternating strategy shows clear superiority.}
\label{tab:strategy}
\end{table}

\subsection{Ablation Study and Comparison}
Here, we evaluate the contributions of each component and framework designs, to further dissect our framework.

\vspace{0.1cm}
\noindent \textbf{Complementarity of pre-training.}
As the detection performance is mainly determined by pseudo-labels, here we show the quality of pseudo-labels, in terms of foreground mIoU and background mIoU. Table~\ref{tab:complementarity} provides the comprehensive results for various settings on both benchmarks.

For common Classification-Based Pre-training, its localization results suffer from the incomplete issue. In detail, the background mIoU is impressive, {\em i.e.} high true negatives, while the foreground mIoU is poor, {\em i.e.} serious false negatives. The main reason is that features pre-trained on action classification datasets only highlight sparse discriminative frames. While the results of Vision-Language Pre-training are just the opposite: suffering from the over-complete issue. The foreground mIoU is considerable, {\em i.e.} high true positives, but the background mIoU is terrible, {\em i.e.} serious false positives. The main reason is that VLP using image-text pairs lacks temporal priors. The above results strongly prove the complementarity between these two pre-training.

Moreover, on both datasets, our method achieves high foreground and background mIoU, that is, more precise and complete localization. This mainly benefits from extensive background supervision provided by the CBP branch, and extensive foreground supervision distilled from the VLP branch. Besides, comparing to VL Pre-training, our results achieve immediate gains on the background mIoU, while only slight drops on the foreground mIoU. This reveals that the lightweight trainable parameters, {\em i.e.} prompts + Transformer, indeed retain the action prior knowledge in VLP, and also significantly suppress false-positive results.

\begin{table}[t]
\small
\centering
\setlength\tabcolsep{7.4pt}
\begin{tabular}[t]{c|cc|ccc|c}
\toprule
\multirow{2}{*}{Model} & \multirow{2}{*}{$\mathcal{L}_{\mathrm{kd}}$} & \multirow{2}{*}{$\mathcal{L}_{\mathrm{fb}}$} & \multicolumn{3}{c|}{mAP@IoU} & \multirow{2}{*}{\begin{tabular}[c]{@{}c@{}}AVG\\ (0.3-0.7)\end{tabular}} \\ \cline{4-6}
&  &  & 0.3 & 0.5 & 0.7 &  \\ \hline \hline
A1 & \checkmark &  & 57.1 & 37.0 & 12.6 & 36.0 \\
A2 &  & \checkmark & 51.7 & 31.2 & 9.4 & 30.9 \\
A3 & \checkmark & \checkmark & \textbf{61.5} & \textbf{42.0} & \textbf{16.8} & \textbf{40.8} \\ 
\bottomrule
\end{tabular}
\vspace{-0.1cm}
\caption{\textbf{Contribution of optimization losses on THUMOS14.} The single knowledge distillation loss $\mathcal{L}_{\mathrm{kd}}$ has brought gratifying localization results, and the additional foreground-background contrastive loss $\mathcal{L}_{\mathrm{fb}}$ further boosts the performance to the best.}
\vspace{0.51cm}
\label{tab:losses}
\end{table}

\begin{table}[t]
\small
\centering
\setlength\tabcolsep{9.5pt}
\begin{tabular}{c|ccc|c}
\toprule
\multirow{2}{*}{Method} & \multicolumn{3}{c|}{mAP@IoU} & \multirow{2}{*}{\begin{tabular}[c]{@{}c@{}}AVG\\ (0.3-0.7)\end{tabular}} \\ \cline{2-4}
 & 0.3 & 0.5 & 0.7 & \\ \hline \hline
UM~\cite{lee2021weakly} & 49.0 & 30.4 & 10.4 & 30.3 \\
UM\,+\,Ours & \textbf{53.9} & \textbf{34.7} & \textbf{13.6} & \textbf{34.3}  \\    \hline
CO2-Net~\cite{hong2021cross} & 54.5 & 38.3 & 13.4 & 35.7 \\
CO2-Net\,+\,Ours & \textbf{56.2} & \textbf{39.7} & \textbf{15.9} & \textbf{37.5} \\
\bottomrule
\end{tabular}
\vspace{-0.1cm}
\caption{\textbf{Framework generalization on THUMOS14.} Our proposed framework is generalized, {\em i.e.} existing methods can serve as our CBP branch, to achieve further improvements.}
\label{tab:tansfer}
\end{table}

\vspace{0.1cm}
\noindent \textbf{Comparison of optimization strategy.}
To evaluate the efficacy of our alternating strategy, Table~\ref{tab:strategy} makes comparison with another two solutions. (1) `Only F step': extract foreground pseudo-labels from vanilla VLP to train the CBP branch; (2) `Only B step': distill background pseudo-labels from the warm-up CBP branch to train the VLP branch. For each solution, we combine the results from two branches via averaging and weighting, respectively.

\begin{figure*}[t]
\begin{center}
\includegraphics [width=0.93\textwidth] {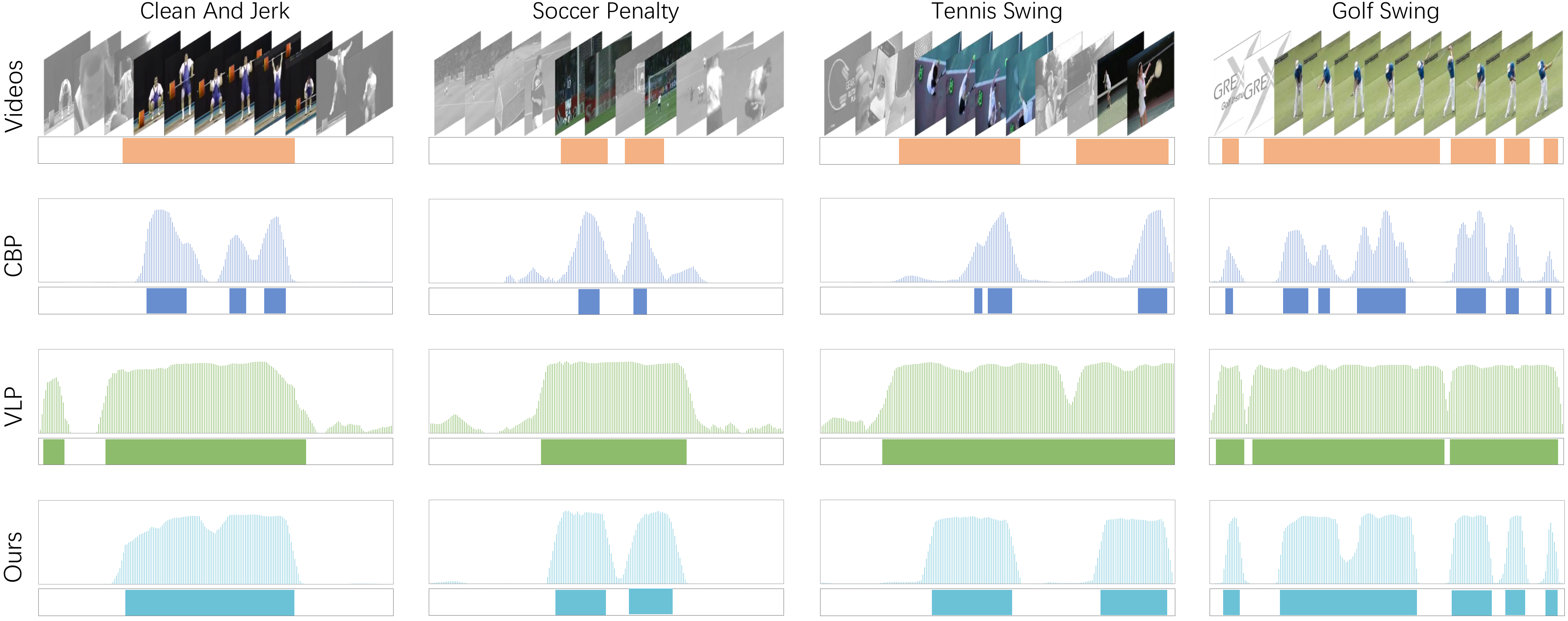}  
\end{center}
\vspace{-0.5cm}
\caption{\textbf{Qualitative comparisons.} The first two rows are videos and ground-truth action intervals. The last six rows are frame-level action probabilities, and localization results of Classification-Based Pre-training (CBP), Vision-Language Pre-training (VLP), and our framework, respectively. CBP suffers from the incomplete issue, while VLP has the over-complete issue. Our framework distills foreground knowledge from VLP and background from CBP, for the strong collaborative alliance, thus bringing more complete and precise results.}
\label{fig:results}
\end{figure*}

`Only F step' performs the worst, mainly suffering from heavy noise in vanilla VLP pseudo-labels (over-complete). By fine-tuning vanilla VLP with many background frames, `Only B step' gets great improvements, but still lacks full fusion of complementarity. Moreover, the weighted operation could suppress noise somewhat, and thus causes better results than the average operation. By comparison, our alternating strategy shows great advantages over competitors, proving the \emph{non-trivial} nature of fusing complementary information, and robust denoising to pseudo-labels.

\vspace{0.1cm}
\noindent \textbf{Contribution of various losses.}
To train our method, both knowledge distillation loss $\mathcal{L}_{\mathrm{kd}}$ and foreground-background contrastive loss $\mathcal{L}_{\mathrm{fb}}$ are leveraged. In Table~\ref{tab:losses}, we analyze the effectiveness of optimization losses on THUMOS14. The single $\mathcal{L}_{\mathrm{kd}}$ (model A1) already brings gratifying performance. This reveals that the confident pseudo-labels of foreground and background distilled from two branches, are well fused by the alternate optimization. On the other hand, with only $\mathcal{L}_{\mathrm{fb}}$, model A2 also performs barely satisfactory results. This is because $\mathcal{L}_{\mathrm{fb}}$ differentiates foreground features and background features, thus eliminating extensive uncertain frames, and also making the localization task easier. Overall, these two losses jointly contribute to the best performance, indicating that both are essential.

\vspace{0.1cm}
\noindent \textbf{Framework generalization.}
Our distillation-collaboration framework is generalized, which means that existing WTAL methods can be employed as the CBP branch. Table~\ref{tab:tansfer} takes two typical methods: CO2-Net~\cite{hong2021cross} and UM~\cite{lee2021weakly}, as examples. Our framework further improves their performance by up to 2-4\% average mAP, showing a good generalization to other methods and other backbone designs.

\subsection{Qualitative Results}
To intuitively demonstrate the superiority, Figure~\ref{fig:results} visualizes detection results from various types of videos.

In general, Classification-Based Pre-training highlights only several discriminative action frames (the incomplete issue), which is more prominent for videos covering low-frequency actions. On the contrary, Vision-Language Pre-training tends to over-activate the action foreground to the background (the over-complete issue), which is especially evident in videos with high-frequency actions. We design the distillation-collaboration framework to fuse the complementarity from these two pre-training. In B step, extensive confident background information is distilled from the well-trained CBP branch, to supervise the VLP branch for false-positive suppression. In F step, sufficient confident foreground locations are distilled from the VLP branch, to guide the CBP branch for false-negative elimination. Hence, our method establishes the strong alliance by collaborative optimization. The detection results are more precise and more complete, regardless of dense or sparse actions.

\section{Conclusion}
This work proposes the novel distillation-collaboration framework to distill free knowledge from Vision-Language Pre-training, for weakly-supervised temporal action localization. Our core insight is that existing VLP often localizes over-complete actions, which is just complementary to the incomplete results of conventional Classification-Based Pre-training. And to form strong alliances, we optimize the framework containing complementary dual branches by an alternating strategy: distill confident background pseudo-labels from the CBP branch, and the confident foreground pseudo-labels from the VLP branch, for collaborative training. Extensive experiments show the significance of distilling VLP and our superior performance. Thorough ablations are studied both quantitatively and qualitatively.

\section{Limitations and Future Work}
For the CBP branch, we freeze the I3D architecture pre-trained on Kinetics~\cite{carreira2017quo}, to extract RGB and Flow features. Such one frozen extractor could save computing resources, but may somewhat limit the performance. 

For the VLP branch, we leverage the CLIP~\cite{Radford21}, which is pre-trained with 400M image-text pairs collected from web, thus could potentially bias towards web data. 

As the future work, we expect more computing resources available, to further optimize our distillation-collaboration framework into the end-to-end training setups, also rendering asynchronous online training.

\vspace{0.3cm}
{\small
\bibliographystyle{ieee_fullname}
\bibliography{egbib}
}

\end{document}